\documentclass[letterpaper]{article} % DO NOT CHANGE THIS
\usepackage[preprint]{aaai2027}  % arXiv preprint mode
\usepackage[hyphens]{url}  % DO NOT CHANGE THIS
\usepackage{graphicx} % DO NOT CHANGE THIS
\usepackage{natbib}  % DO NOT CHANGE THIS AND DO NOT ADD ANY OPTIONS TO IT
\usepackage{caption} % DO NOT CHANGE THIS AND DO NOT ADD ANY OPTIONS TO IT
\usepackage{amsmath}
\usepackage{algorithm}
\usepackage{algorithmic}
\usepackage[most]{tcolorbox}      
\usepackage{enumitem}          
\usepackage{calc}                     
\usepackage{xcolor}  
\usepackage{listings}
\tcbuselibrary{breakable,listings,skins}
\newcommand{\method}{Project2Task}

\title{\method{}: Graph-Guided Project-Level Planning for \\Autonomous Research}
\author{
    Huirui Xu\textsuperscript{\rm 1,\rm 2},
    Runtao Xu\textsuperscript{\rm 1,\rm 3},
    Shuo Ren\textsuperscript{\rm 1}\corresponding,
    Jiajun Zhang\textsuperscript{\rm 1,\rm 2,\rm 4}\corresponding
}

\affiliations{
    \textsuperscript{\rm 1}Institute of Automation, Chinese Academy of Sciences\\
    \textsuperscript{\rm 2}School of Artificial Intelligence, University of Chinese Academy of Sciences\\
    \textsuperscript{\rm 3}School of Future Technology, University of Chinese Academy of Sciences  \textsuperscript{\rm 4}Wuhan AI Research\\
    \{xuhuirui2025,xuruntao2026,shuo.ren\}@ia.ac.cn, jjzhang@nlpr.ia.ac.cn\\
}
\begin{document}
\maketitle

\begin{abstract}
Research agents are increasingly able to search the literature, propose hypotheses, generate code, run experiments, and draft manuscripts from a single topic. Yet a research project is not simply a larger task: it is a long-horizon research agenda that must be advanced through multiple bounded tasks with distinct but related research objectives. These tasks may explore alternative approaches to the project in parallel or address successive challenges in a dependency-aware sequence. This creates a planning problem that current single-task systems handle poorly. They either assign the entire project as one oversized task, split it into a flat list of vague and overlapping tasks, or leave humans to manually coordinate task boundaries and execution order.
We introduce \method{}, a project-level planning layer for autonomous research systems. Given a macro project brief, \method{} constructs an innovation-atom lineage graph, selects a portfolio decomposition strategy using a lightweight Bernoulli block-model objective, generates bounded autoresearch tasks with explicit contribution ownership, resolves overlaps and fills in missing execution fields, and emits dependency-aware execution contracts that are independent of any particular downstream research executor. Each task is a bounded, executable research unit with its own objectives, inputs, expected artifacts, evaluation requirements, boundary constraints, and dependencies; together, the tasks advance the overarching project. Task artifacts may include code, datasets, benchmarks, experimental results, analyses, reports, or manuscripts.
% Rather than reporting final experimental results, we specify an evaluation protocol for project-level autonomous research. This protocol constructs datasets from macro-project briefs and evaluates the generated paper portfolios across five core dimensions—coherence, content overlap, macro-goal coverage, content consistency, and task division rationality—by comparing planning-based decomposition against naive baselines. The central claim is that autonomous research systems need an explicit project-to-paper planning contract before single-paper execution can be reliable.
On a benchmark of ten project briefs yielding roughly 30 tasks, a manuscript-based portfolio evaluation shows that Project2Task achieves an average portfolio-quality score of 7.15, compared with 4.58 for the Brief Baseline and 5.31 for the Topic-only Setting. Its structured contracts also increase AutoResearchClaw’s downstream task accuracy from 0.536 to 0.759. These results demonstrate the value of an explicit project-to-task planning layer for producing coherent, non-redundant, and executable portfolios of autoresearch tasks.
\end{abstract}

\section{Introduction}

Autonomous research agents are increasingly capable of performing literature retrieval, hypothesis generation, method design, implementation, experimentation, and result analysis \citep{lu2024aiscientist,schmidgall2025agentlaboratory,liu2026autoresearchclawselfreinforcingautonomousresearch}. Most existing systems organize these capabilities around a single research task. Typically, a research task is a bounded research problem with a specific objective, a clear scope, and a distinct contribution. It can be independently tested and executed end-to-end by an agent to produce the corresponding artifacts.

However, in practice, research is often organized around projects that extend beyond the scope of a single task. A research project is a broader, long-horizon agenda that can typically be decomposed into multiple well-defined research tasks whose organization reflects the underlying research logic. Tasks exploring different approaches to the same challenge may proceed in parallel, whereas tasks addressing successive challenges may form a sequence in which later tasks build on earlier outcomes. For example, a doctoral research project on reliable autonomous agents may involve separate tasks concerning persistent memory, adaptive planning, and result verification. Persistent memory may provide a shared foundation for planning and verification, allowing the latter two tasks to proceed in parallel. Although current autonomous research agents are effective at executing well-scoped tasks, applying them directly to broader projects without explicit planning can lead to overlapping tasks, inconsistent assumptions, and missing dependencies. Consequently, individually plausible outputs may fail to form a coherent project-level result.

We formalize this missing layer as \emph{project-to-task planning} for autonomous research, and propose \textsc{Project2Task}, a graph-guided planning mechanism that transforms a broad project brief into multiple executable tasks with clear objectives, bounded scopes, and explicit contribution ownership. \textsc{Project2Task} represents candidate contributions as innovation atoms and organizes them in a directed lineage graph. It selects among horizontal, vertical, and hybrid decomposition strategies using a lightweight Bernoulli block-model objective. Among multiple plausible decompositions, it seeks the best one that preserves project coverage while maintaining clear task boundaries, distinct contribution ownership, and valid dependencies. It then synthesizes and repairs the resulting tasks to enforce portfolio-level coherence. The final output consists of executor-agnostic task contracts that specify each task's objective, scope, contribution ownership, shared assets, evaluation requirements, dependencies, and execution order. These contracts provide a structured interface for downstream research agents to execute tasks independently while enabling their outputs to be integrated into a coherent project-level result.

To evaluate \textsc{Project2Task}, we design a project-level autonomous-research evaluation protocol based on a benchmark of ten project-level research briefs. The protocol uses generated manuscripts as downstream artifacts and assesses task-portfolio quality along five dimensions: coherence, coverage, overlap control, consistency, and task division. \textsc{Project2Task} achieves an average portfolio-quality score of $7.15$, compared with $4.58$ for the Brief Baseline and $5.31$ for the Topic-only Setting. Replacing the selected decomposition strategy with the second-ranked one reduces the score to $6.48$, demonstrating the importance of topology-aware decomposition. Integrating \textsc{Project2Task} contracts with AutoResearchClaw increases average downstream task accuracy from $0.536$ to $0.759$.

\begin{figure*}[t]
\centering
\includegraphics[width=\textwidth]{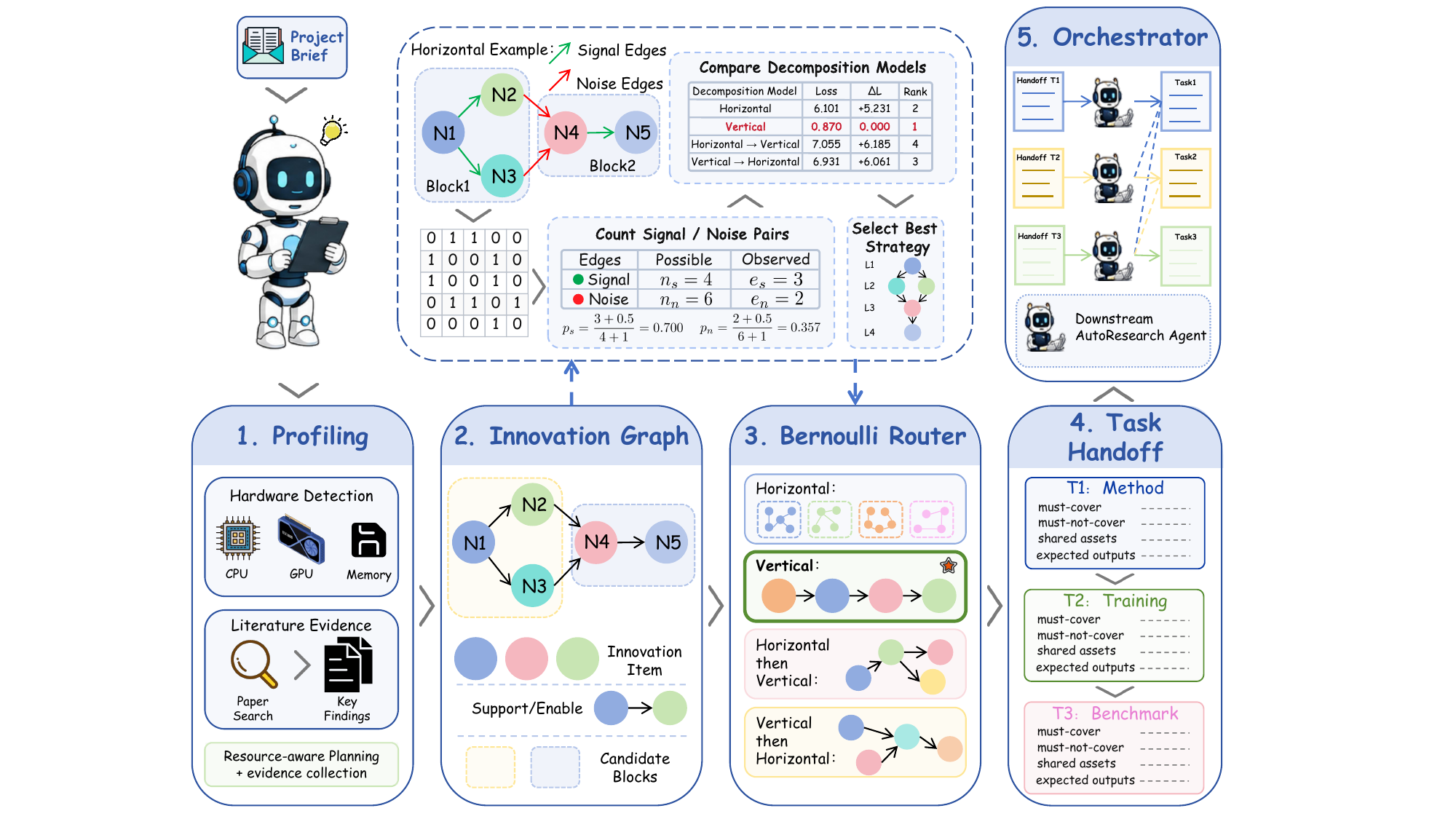}
\caption{\method{} converts a macro research project into an executable task portfolio. The planner first profiles the project brief with available code, resources, and literature evidence, then decomposes the project into innovation atoms and builds a lineage graph whose edges store endpoint pairs and rationales describing dependency-like relations. A Bernoulli block-model router compares horizontal, vertical, and hybrid decomposition strategies on this graph, after which the selected plan is repaired into task-level execution contracts specifying contribution ownership, shared assets, must-cover and must-not-cover content, expected outputs, and dependency order. The orchestrator uses these contracts to coordinate task executors, making the project structure explicit before downstream task execution begins.}
\label{fig:planner-overview}
\end{figure*}

Figure~\ref{fig:planner-overview} summarizes the complete planning and execution pipeline.

Our contributions are as follows:
\begin{enumerate}
    \item We formalize project-to-task planning for autonomous research, in which a broad project brief is transformed into a configurable portfolio of bounded tasks with explicit objectives, boundaries, contribution ownership, shared assets, dependencies, and execution order.
    \item We propose \textsc{Project2Task}, a planner that operationalizes this problem through innovation atoms, a lineage graph, decomposition routing, plan repair, dependency scheduling, and structured task contracts for downstream executors.
    \item We design a project-level dataset and evaluation protocol that uses generated manuscripts as downstream artifacts to assess task-portfolio quality along five dimensions: coherence, coverage, overlap control, consistency, and task division.
\end{enumerate}
\section{Related Work}
\label{sec:related}

\paragraph{Autonomous research agents.} Language agents are increasingly used to automate scientific workflows. The AI Scientist generates ideas, edits code, runs experiments, writes papers, and performs automated review \citep{lu2024aiscientist}, while Agent Laboratory organizes research assistance into literature review, experimentation, and report writing \citep{schmidgall2025agentlaboratory}. A closely related autoresearch line treats a research codebase and its measurements as the object of iterative improvement \citep{ferreira2026autoresearch}: compact codebases such as nanoGPT provide practical substrates for closed-loop experimentation \citep{karpathy2022nanogpt}; AutoResearchClaw studies self-reinforcing research with multi-agent feedback, execution repair, verifiable reporting, and human-AI collaboration \citep{liu2026autoresearchclawselfreinforcingautonomousresearch}; and Bilevel Autoresearch meta-optimizes the research loop itself \citep{qu2026bilevel}. These systems primarily operate around one research objective, whereas \method{} addresses the upstream planning problem of turning a broad project into a coordinated portfolio of tasks.

\paragraph{Scientific writing and literature synthesis.} Retrieval-augmented systems ground scientific writing in external evidence. PaperQA answers questions over full-text scientific papers \citep{lala2023paperqa}, OpenScholar scales citation-backed literature synthesis with a large scientific datastore \citep{asai2024openscholar}, and STORM constructs outlines through retrieval and multi-perspective question asking before drafting long-form articles \citep{shao2024storm}. More recent systems move toward interactive and content-grounded literature analysis: InsightAgent uses multiple agents to support human-guided systematic reviews, while IntrAgent uses iterative, content-grounded retrieval and reading \citep{qiu-etal-2025-completing,ma-etal-2026-intragent}. Our focus is different: \method{} plans contribution boundaries and dependencies before downstream task execution begins, instead of optimizing evidence collection, outlining, or single-document synthesis.

\paragraph{LLM planning and multi-agent coordination.} General agent methods decompose tasks and coordinate action. ReAct interleaves reasoning with environment interaction \citep{yao2022react}, Plan-and-Solve separates plan generation from execution \citep{wang2023planandsolve}, and AutoGen, CAMEL, and MetaGPT provide conversational or role-based multi-agent collaboration frameworks \citep{wu2023autogen,li2023camel,hong2023metagpt}. Recent work also studies graph-structured agent workflows and orchestration: AFlow searches over workflows represented as code, while OrchestrationBench evaluates agent orchestration under constraints in scenarios involving sequential and parallel tool use \citep{zhang2025aflow,ahn2026orchestrationbench}. These methods focus on how given tasks are planned and executed, whereas project-to-task planning determines which bounded research tasks a project should contain and how they relate before execution.

\paragraph{Evaluation of research agents.} Recent benchmarks evaluate whether agents can perform scientific or engineering tasks, including machine-learning experimentation in MLAgentBench \citep{huang2023mlagentbench}, data-driven scientific tasks in ScienceAgentBench \citep{chen2024scienceagentbench}, and paper replication in PaperBench \citep{starace2025paperbench}. Recent benchmarks also move toward more open-ended and end-to-end research settings: MLR-Bench evaluates research systems both at individual stages and end to end, while ResearchGym provides executable environments for studying closed-loop research processes \citep{chen2025mlrbench,garikaparthi2026researchgym}. These benchmarks evaluate agents after research objectives are specified, whereas our evaluation asks whether a planner first constructs a coherent project-level task portfolio.

\section{Method}
\label{sec:method}

\subsection{Portfolio Planning Formulation}
We define \emph{project-to-task planning} as transforming a macro research project into a dependency-aware portfolio of bounded autoresearch tasks. Each task is a bounded, executable research unit within the project, with explicit objectives, inputs, expected artifacts, evaluation requirements, boundary constraints, and dependencies. Task artifacts may include code, datasets, benchmarks, experimental results, analyses, reports, or manuscripts. The input is $X=(B,R,L)$, where $B$ is a structured project brief, $R$ is a resource profile, and $L$ is literature evidence. These objects provide the project goal, candidate ideas, constraints, feasible validation scale, and external evidence for candidate claims.

The planner builds a directed acyclic lineage graph (DAG) $G=(V,E_G)$. We assume that project dependencies are acyclic so that the resulting tasks admit an executable order. Source nodes are assigned topological level $0$, and every other node is assigned one plus the maximum level of its predecessors. Each node $v\in V$ is an \emph{innovation atom} with a title, description, problem to solve, innovation claim, validation path, and evidence. Each edge stores only source, target, rationale, and optional required artifact. The lineage prompt asks edges to capture dependency-style relations such as support, constraint, or motivation.

The output is $\Pi=(T,D,H)$. $T=\{t_1,\ldots,t_k\}$ is a set of tasks, $D$ is a dependency model, and $H$ is a set of per-task contracts. The portfolio size $k$ is a planning-granularity parameter. It may be supplied by the user or estimated from project scope, graph structure, resources, and validation signals. Each task owns a scoped contribution and records assets, evaluation plans, non-goals, must-cover and must-not-cover constraints, source atoms, and sibling boundary rules.

A valid portfolio is therefore more than a list of titles. It should cover the macro project goal, assign core claims to task-specific owners, make reusable assets explicit, preserve feasible validation paths, and expose dependencies between tasks. These constraints are important because the same project code, dataset, or evaluation protocol may be shared across tasks, while the primary objective and output ownership of each task must remain distinct.

\subsection{Project-to-Graph Construction}
This stage constructs the planning state used by the router. It does not decide task boundaries. Instead, it prepares a structured brief, resource constraints, literature evidence, and a lineage graph whose nodes are finer-grained than complete tasks.

\textbf{Brief, resource, and evidence grounding.}
% The parser turns the raw request into a structured brief that records the macro goal, target problem, candidate ideas, constraints, expected deliverables, and evaluation expectations. The resource profiler records available hardware and runtime constraints. The search component retrieves project-level literature evidence. These objects condition the graph-building prompt and later planning stages.
The parser turns the raw request into a structured brief that records the macro goal, target problem, candidate ideas, constraints, expected deliverables, and evaluation expectations. The planner then builds a resource profile $R$ from the available environment and user constraints, recording available hardware information and using it to ensure that planned evaluations remain feasible under the available resources. Given $B$ and $R$, the planner retrieves literature evidence $L$ through project-specific queries. The retrieved papers, trends, and gaps are then used to ground candidate atoms and provide literature focus for later task contracts.

\textbf{Lineage graph construction.}
The lineage builder then produces $G=(V,E_G)$. Each node follows the schema above and is intended to be finer-grained than a complete task. A node may describe a method module, dataset construction step, benchmark protocol, or validation mechanism. Edges connect atoms that should be considered together for planning, using the prompt-level dependency semantics above. The graph object also records a project summary, gap distribution, validation story, and resource notes. The resulting graph is not a complete task plan. It is a structured planning object from which the router can estimate modularity, dependency depth, foundation structure, and validation independence. 

\subsection{Graph-Guided Decomposition Routing}
The router compares four decomposition templates on the same lineage graph. We score these templates with a Bernoulli block-model view inspired by stochastic block models \citep{holland1983stochastic,karrer2011stochastic}. The analogy is limited to the likelihood calculation: \method{} does not infer latent communities from data. Instead, each strategy $s$ constructs a candidate block assignment $b_s:V\rightarrow\mathcal{B}_s$ from the lineage graph and labels template-induced dyad observations as signal or noise. Signal observations are where the template expects an edge; noise observations are where it does not. Directional templates also check the expected edge direction.

\textbf{Horizontal.}
The horizontal template clusters atoms into greedy modularity communities on the undirected projection of $G$. Dyads within a community are signal positions, and dyads across communities are noise positions. This template matches projects with separable contribution modules.

\textbf{Vertical.}
The vertical template assigns atoms to topological levels. Forward dyads from level $\ell$ to $\ell+1$ are signal positions; same-level, skipped-level, and backward relations are noise. This template matches staged research chains.

\textbf{Horizontal-then-vertical.}
This template first clusters non-sink atoms into horizontal communities, then assigns a selected sink atom to a special \textsc{SINK} block. Signal positions include within-community dyads and directed edges from upstream blocks to the sink. This captures parallel modules followed by an integration task.

\textbf{Vertical-then-horizontal.}
This template assigns a foundation hub atom to a \textsc{FOUNDATION} block and clusters the remaining atoms into branch communities. Signal positions include edges from the foundation to branches and dyads within each branch. This captures a shared foundation followed by specialized tasks.

\textbf{Template feasibility.}
Let $\mathcal{S}$ denote the four decomposition templates described above. For a lineage graph $G$, let $\mathcal{S}_G\subseteq\mathcal{S}$ denote the templates whose block assignments can be constructed on $G$. Horizontal and vertical templates are always feasible under the DAG assumption. A hybrid template is included only when its required \textsc{SINK} or \textsc{FOUNDATION} block can be formed; otherwise, it is excluded before NLL ranking.

For scoring, $G$ is converted into a binary adjacency matrix $A$, where $A_{ij}=1$ iff $(v_i,v_j)\in E_G$. Given $b_s$, the template induces an observation set $\mathcal{O}_s$. Each observation $o=(i,j,r_o)$ records a dyad or oriented dyad and its role $r_o\in\{+,-\}$, where $+$ denotes a signal position and $-$ denotes a noise position. The block counts are
\[
n_s^{r}=\sum_{o\in\mathcal{O}_s}\mathbf{1}[r_o=r],\qquad
e_s^{r}=\sum_{o=(i,j,r_o)\in\mathcal{O}_s}\mathbf{1}[r_o=r]A_{ij},
\]
for $r\in\{+,-\}$. These counts instantiate a two-parameter Bernoulli block model for each template. The smoothed estimates are
\[
\hat{p}_s^{+}=\frac{e_s^{+}+0.5}{n_s^{+}+1.0},\qquad
\hat{p}_s^{-}=\frac{e_s^{-}+0.5}{n_s^{-}+1.0}.
\]

We define
\begin{equation}
\ell(e,n,p)=-e\log p-(n-e)\log(1-p).
\end{equation}
The strategy score is
\begin{equation}
\mathrm{NLL}(s)=
\ell(e_s^{+}, n_s^{+}, \hat{p}_s^{+})
+
\ell(e_s^{-}, n_s^{-}, \hat{p}_s^{-}).
\label{eq:router-nll}
\end{equation}
The selected strategy is
\[
s^\star=\arg\min_{s\in\mathcal{S}_G}\mathrm{NLL}(s).
\]

Algorithm~\ref{alg:decomposition-routing} summarizes the procedure. The selected strategy is returned together with confidence, rationale, strategy guidance, rejected strategies, and graph-derived task blueprints.

\begin{algorithm}[t]
\caption{Graph-guided decomposition routing}
\label{alg:decomposition-routing}
\begin{algorithmic}[1]
\REQUIRE Lineage graph $G=(V,E_G)$; strategy set $\mathcal{S}$
\ENSURE Selected strategy $s^\star$, blocks $b_{s^\star}$, diagnostics
\STATE Compute graph diagnostics from $G$
\STATE Form $\mathcal{S}_G$ by excluding infeasible hybrid templates
\FOR{each $s \in \mathcal{S}_G$}
    \STATE Construct template-specific block assignment $b_s$
    \STATE Derive signal/noise dyads from $b_s$
    \STATE Estimate $\hat{p}_s^{+}$ and $\hat{p}_s^{-}$ with smoothing
    \STATE Compute $\mathrm{NLL}(s)$ under Eq.~\ref{eq:router-nll}
\ENDFOR
\STATE Select $s^\star=\arg\min_{s\in\mathcal{S}_G}\mathrm{NLL}(s)$
\STATE \textbf{return} $s^\star$, rejected strategies, and task blueprints
\end{algorithmic}
\end{algorithm}

Given the template-specific block assignments, likelihood scoring scans the dyads once per strategy and costs $O(|\mathcal{S}||V|^2)$.

\begin{table*}[t]
\centering
\small
\renewcommand{\arraystretch}{1.3} 
\begin{tabular}{p{0.22\linewidth}|p{0.34\linewidth}|p{0.34\linewidth}}
\hline
\textbf{Dimension} & \textbf{Question} & \textbf{Evidence} \\
\hline
Coherence & Do the papers form a logically connected and cohesive project narrative? & Logical relationships between papers, unified research objectives, cross references and complementary content. \\
\hline
Overlap Control & Do the papers keep their core ideas, methods, and claims distinct while limiting redundant repetition? & Text and core claim repetition, distinction between foundational content and innovative contributions. \\
\hline
Coverage & Does the paper collection fully cover the core research topics and key scientific questions? & Key research dimensions, technical points and application scenarios covered by all papers. \\
\hline
Consistency & Are terminology, assumptions, data and experimental settings consistent across all papers? & Unified definitions, theoretical preconditions, data processing rules and evaluation metrics. \\
\hline
Task Division & Is the decomposition of the overall research topic into tasks reasonable and balanced? & Task-division logic, clear boundaries, workload distribution, and allocation of core innovations. \\
\hline
\end{tabular}
\caption{Dimensions used to assess task-portfolio quality from generated manuscripts.}
\label{tab:metrics}
\end{table*}

\subsection{Task Contract Generation}
After routing, the planner turns graph-level blueprints into a portfolio of bounded task specifications. This stage connects the selected decomposition structure with executable tasks.

\textbf{Task synthesis.}
The planner decides $k$ and generates $T=\{t_1,\ldots,t_k\}$ from the brief, evidence, resource profile, lineage graph, and routing decision. Router blocks are used as blueprints, not hard partitions. The planner may merge or split graph blocks for a task when doing so yields clearer contribution ownership and a feasible evaluation plan. The resulting task plan records tasks, required portfolio content, forbidden overlap zones, shared assets, an overall evaluation plan, portfolio success criteria, risks, and open questions.

\textbf{Validation and repair.}
The portfolio is checked before handoff. The evaluator measures coverage, claim overlap, literature grounding, coherence, dependency feasibility, and execution readiness. Blocking issues include missing core contributions, missing evaluation plans, target-count mismatch, high-risk claim overlap, and missing sibling boundaries. A bounded repair loop may tighten scopes, separate overlapping claims, fill missing evaluation fields, add expected outputs, or write boundary rules between sibling tasks. This step checks contract executability, not final output quality.

\textbf{Contract export.}
The repaired portfolio is converted into a dependency model and per-task contracts. The dependency model contains edges, execution order, parallel groups, critical path, and roadmap. Dependency edges may specify upstream concepts, artifacts, methods, datasets, evaluations, or other required deliverables. Each contract contains the task topic, scoped goal, project context, role in the portfolio, boundaries, shared assets, dependencies, literature focus, experiment direction, evaluation guidance, reporting guidance, expected outputs, and success criteria.

\subsection{Downstream Task Integration}
\method{} exposes a generic adapter interface. A downstream adapter selects a per-task contract, renders it as the executor's task input, and keeps portfolio boundaries available during execution. The adapter does not need to change the planner, because the contract already contains the task scope and dependency context.

The same contract can be used as persistent context, stage-level guidance, or checklist constraints. For stage-based executors, scoping, literature, experiment, implementation, analysis, drafting, and review guidance can be projected to the corresponding workflow stages. For non-stage-based executors, the fields can be used as task constraints, tool policies, or review criteria. The key requirement is that the portfolio boundary remains visible throughout execution, rather than appearing only in the initial prompt.

The dependency model coordinates multiple tasks. Independent tasks may run in parallel, while dependent tasks wait for upstream artifacts such as datasets, code, benchmarks, definitions, or evaluation results. During and after execution, the original contract also provides an audit reference for boundary drift, missing evaluation evidence, duplicated contribution claims, or violations of must-cover and must-not-cover constraints.

\section{Experiments}
\label{sec:experiments}

\subsection{Experiment Setup}
\label{sec:experimental_setup}

\subsubsection{Dataset:}
Project-level autonomous research should be evaluated at the portfolio level rather than at the level of a single task alone. Each benchmark instance is therefore represented as a macro research project with six abstract dimensions: research theme, background, core problem, methodology, expected outcomes, and research significance.
The constructed project-level research dataset is derived from two established academic benchmark sources with standardized curation and augmentation procedures. Specifically, we select 7 representative single-task samples from the NanoResearch dataset \citep{xu2026nanoresearchcoevolvingskillsmemory} and 3 valid samples from the ARC-Bench dataset \citep{liu2026autoresearchclawselfreinforcingautonomousresearch} to form the base data pool. The resulting 10 project instances yield roughly 30 tasks and span several representative research domains, including Computer Vision (CV), Natural Language Processing (NLP), tabular machine learning, time series analysis, graph representation learning, audio processing, and multimodal learning. This domain diversity allows us to test the planner across different project settings, while the limited scale means the results should be interpreted as an initial project-level evaluation rather than as evidence of broad generalization. Unlike fine-grained, task-specific records in single-paper benchmarks, all selected base samples are generalized and upgraded into complete project-level research instances via large-language-model-driven augmentation. Detailed dataset construction methods are provided in Appendix~\ref{app:prompt_template}.

\subsubsection{Evaluation protocol:}
Table~\ref{tab:metrics} summarizes the evaluation metrics. We evaluate task-portfolio quality using the collection of generated manuscripts along five core dimensions: coherence, overlap control, coverage, consistency, and task division. For each dimension, an LLM judge outputs a score from 1 to 10 along with a brief justification based on the project theme and the full texts of the generated papers. This experiment therefore operationalizes task-portfolio quality through manuscript outputs; other autoresearch artifacts are outside its evaluation scope. We use Qwen3.6-Plus as the judge model \citep{qwen36plus} and keep the judge prompt, input format, scoring scale, and sampling settings identical across all compared methods. We average each reported score across five independent judge runs. In each run, we provide the judge with identical portfolio evidence and ask it to assign a score together with a concise justification. Full evaluation scoring criteria are provided in the Appendix~\ref{app:evaluation_protocol}.

\begin{figure}[t!]
    \centering
    \includegraphics[width=\linewidth]{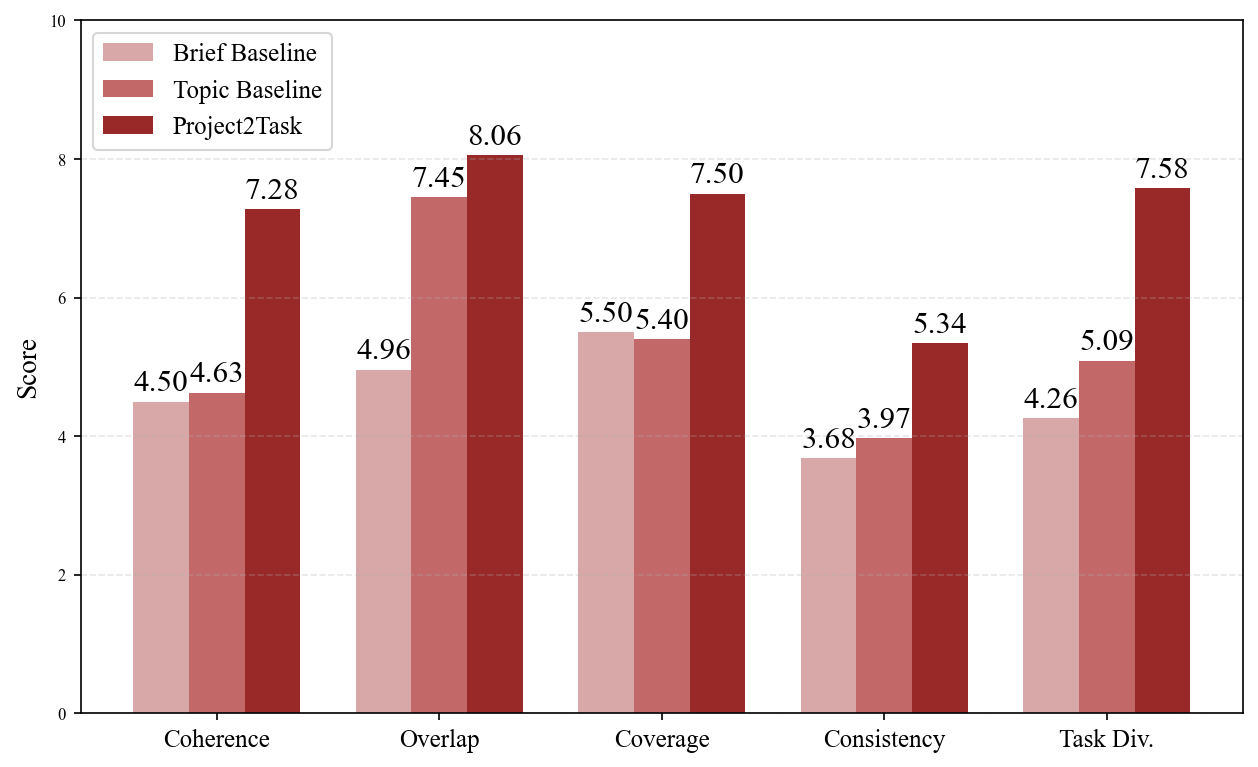}
    \caption{Task-portfolio quality across planning methods, evaluated using generated manuscripts. All metrics are scored on a scale from 1 to 10, where higher scores indicate better performance across all dimensions.}
    \label{fig:main_results}
\end{figure}

\subsubsection{Compared settings:}
We design three sets of experiments to evaluate our method: (1) portfolio-quality comparison against alternative prompting settings, (2) ablation on the graph-guided strategy routing mechanism, and (3) downstream execution evaluation on concrete research tasks.
\textbf{Portfolio comparison.} We compare our project-level planner (\textit{Project2Task}) with two configurations of the original AutoResearchClaw pipeline. In the \textbf{Brief Baseline}, the overarching project topic is fed directly into the original AutoResearchClaw model to generate papers in isolation. In the \textbf{Topic-only Setting}, we use the same task topics generated by our planner but remove all structured planning information; only concise task titles are extracted and fed into the model. This setting isolates the value of the structured task contracts beyond task-topic decomposition alone.
\textbf{Routing ablation.} A core technical contribution of our method is the Bernoulli block-model-based graph routing mechanism, which selects a decomposition strategy according to the innovation-atom lineage graph and the routing objective. To examine the contribution of this strategy-selection step, we construct a variant denoted as \textit{Runner-up}. All pipeline components remain identical to the main experiment, except that the router is forced to use the second-ranked decomposition strategy (e.g., Horizontal, Vertical, Horizontal then Vertical, or Vertical then Horizontal) instead of the selected one. This isolates the impact of graph-guided strategy routing.
\textbf{Downstream execution.} To examine whether our planner affects concrete task execution, we transform the original 10 project queries into executable research tasks by specifying: (1) a precise problem statement, (2) designated datasets, (3) baseline methods for comparison, and (4) explicit optimization metrics. The baseline is the original AutoResearchClaw system receiving only a topic description. Our method first applies project-level planning to decompose the task, then feeds the structured handoff contracts into AutoResearchClaw for execution. The reported metric is task accuracy, defined as the test-set performance of the methods autonomously designed and implemented by the research agent. Because these tasks span different datasets and domains, the averaged accuracy is used as a coarse summary; the task-level results remain the primary evidence.

\begin{figure}[t!]
    \centering
    \includegraphics[width=\linewidth]{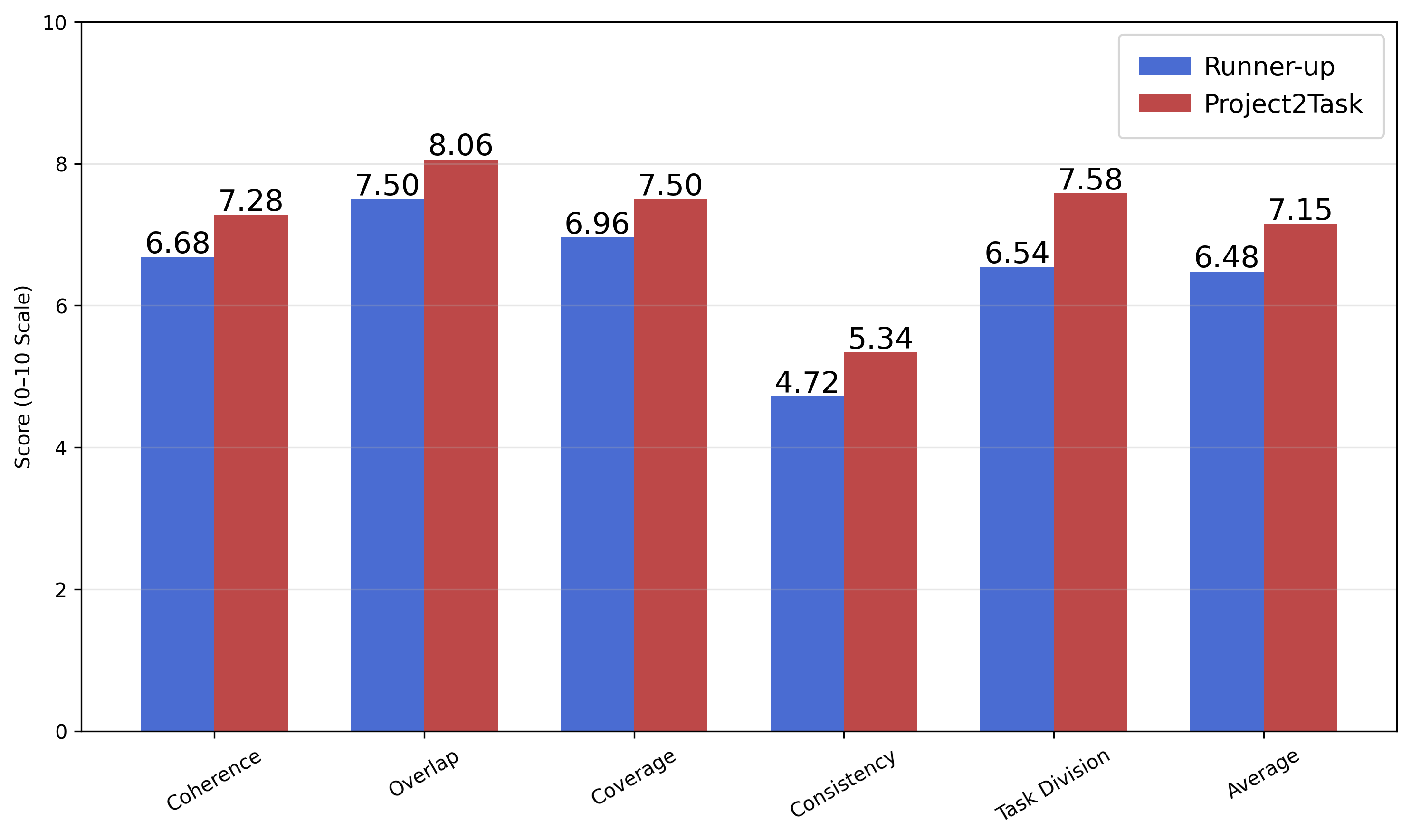}
    \caption{Overall portfolio quality comparison between the selected strategy (\textit{Project2Task}) and the second-best strategy (\textit{Runner-up}). Higher scores indicate better performance.}
    \label{fig:overall_ablation}
\end{figure}

\subsection{Main Results}
\label{sec:main_results}

\subsubsection{Portfolio Quality:}
In this section, we present the main evaluation results of our proposed project-level planner compared against the two baseline settings. The quantitative numerical results are summarized in Figure~\ref{fig:main_results}.

As shown in Figure~\ref{fig:main_results}, our proposed \textit{Project2Task} method obtains higher scores than both comparison settings across all evaluation dimensions, with an average score of 7.15 compared with 5.31 for \textit{Topic} and 4.58 for \textit{Brief}. All reported scores are the means of five independent evaluation runs, and the detailed query-level results with standard deviations are provided in Table~\ref{tab:detailed_scores}.

\begin{table*}[t]
\centering
\small
\begin{tabular}{c|ccc}
\hline
\textbf{Query} & \textbf{Brief (Mean ± SD)} & \textbf{Topic (Mean ± SD)} & \textbf{Project2Task (Mean ± SD)} \\
\hline
1 & $4.96 \pm \text{\scriptsize 0.39}$ & $5.80 \pm \text{\scriptsize 0.42}$ & $6.68 \pm \text{\scriptsize 0.39}$ \\
2 & $5.36 \pm \text{\scriptsize 0.23}$ & $5.44 \pm \text{\scriptsize 0.39}$ & $7.56 \pm \text{\scriptsize 0.42}$ \\
3 & $2.68 \pm \text{\scriptsize 0.24}$ & $6.28 \pm \text{\scriptsize 0.16}$ & $6.64 \pm \text{\scriptsize 0.36}$ \\
4 & $5.96 \pm \text{\scriptsize 0.41}$ & $5.00 \pm \text{\scriptsize 0.13}$ & $6.96 \pm \text{\scriptsize 0.32}$ \\
5 & $4.40 \pm \text{\scriptsize 0.13}$ & $5.24 \pm \text{\scriptsize 0.45}$ & $7.32 \pm \text{\scriptsize 0.37}$ \\
6 & $3.96 \pm \text{\scriptsize 0.15}$ & $4.80 \pm \text{\scriptsize 0.36}$ & $7.16 \pm \text{\scriptsize 0.28}$ \\
7 & $3.80 \pm \text{\scriptsize 0.36}$ & $4.20 \pm \text{\scriptsize 0.25}$ & $6.84 \pm \text{\scriptsize 0.39}$ \\
8 & $5.32 \pm \text{\scriptsize 0.27}$ & $5.00 \pm \text{\scriptsize 0.28}$ & $7.40 \pm \text{\scriptsize 0.44}$ \\
9 & $4.24 \pm \text{\scriptsize 0.27}$ & $6.36 \pm \text{\scriptsize 0.15}$ & $7.52 \pm \text{\scriptsize 0.42}$ \\
10 & $5.12 \pm \text{\scriptsize 0.20}$ & $4.96 \pm \text{\scriptsize 0.29}$ & $7.44 \pm \text{\scriptsize 0.26}$ \\
\hline
\textbf{Average} & \textbf{4.58} & \textbf{5.31} & \textbf{7.15} \\
\hline
\end{tabular}
\caption{Detailed evaluation scores (mean ± standard deviation) for each query under various planning methods; means are computed over five repeated evaluations.}
\label{tab:detailed_scores}
\end{table*}

\textbf{Coherence and Coverage.} We observe the largest gains in Coherence (7.28 vs. 4.63/4.50) and Coverage (7.50 vs. 5.50/5.40). These results suggest that project-level planning helps produce a more connected research narrative and broader coverage of the project's core scientific questions.
\textbf{Overlap Control and Consistency.} Our method obtains the highest Overlap Control score (8.06). This suggests that decomposition helps distinguish shared background from task-specific contribution claims in the generated manuscripts, reducing redundant arguments across the portfolio. The improved Consistency score (5.34) also indicates more unified terminology, assumptions, and experimental settings.
\textbf{Task Division Rationality.} The Task Division score reaches 7.58, higher than both comparison settings. This suggests that the planner can split a macro project into more clearly bounded tasks while keeping each task focused on a distinct research role.

Overall, the results indicate that \method{} improves project-level portfolio quality over the brief-only and topic-only settings in this evaluation.

\subsubsection{Routing Ablation:}

The quantitative results are summarized in Figure~\ref{fig:overall_ablation}. Compared with the runner-up decomposition strategy, the strategy selected by our graph-guided router improves the average portfolio-quality score from 6.48 to 7.15. The largest gain appears in \textbf{Task Division} (+1.04), suggesting that matching the decomposition structure to the lineage graph helps produce clearer task boundaries and more balanced task allocation. The improvements in \textbf{Overlap Control} (+0.56) and \textbf{Consistency} (+0.62) further indicate that the selected strategy reduces redundant novelty claims and helps maintain shared assumptions across the portfolio.

\begin{center}
\small
\begin{tabular}{@{}ccc@{}}
\hline
\textbf{Query} & \textbf{Runner-Up} & \textbf{Project2Task} \\
\hline
1  & $\textbf{6.84}\pm\text{\scriptsize 0.32}$ & $6.68\pm\text{\scriptsize 0.39}$ \\
2  & $6.92\pm\text{\scriptsize 0.37}$ & $\textbf{7.56}\pm\text{\scriptsize 0.42}$ \\
3  & $5.96\pm\text{\scriptsize 0.34}$ & $\textbf{6.64}\pm\text{\scriptsize 0.36}$ \\
4  & $6.48\pm\text{\scriptsize 0.30}$ & $\textbf{6.96}\pm\text{\scriptsize 0.32}$ \\
5  & $6.52\pm\text{\scriptsize 0.39}$ & $\textbf{7.32}\pm\text{\scriptsize 0.37}$ \\
6  & $6.08\pm\text{\scriptsize 0.20}$ & $\textbf{7.16}\pm\text{\scriptsize 0.28}$ \\
7  & $6.16\pm\text{\scriptsize 0.34}$ & $\textbf{6.84}\pm\text{\scriptsize 0.39}$ \\
8  & $6.52\pm\text{\scriptsize 0.20}$ & $\textbf{7.40}\pm\text{\scriptsize 0.44}$ \\
9  & $6.20\pm\text{\scriptsize 0.33}$ & $\textbf{7.52}\pm\text{\scriptsize 0.42}$ \\
10 & $7.12\pm\text{\scriptsize 0.16}$ & $\textbf{7.44}\pm\text{\scriptsize 0.26}$ \\
\hline
\textbf{Average}  & \textbf{6.48} & \textbf{7.15}\\
\hline
\end{tabular}
\captionof{table}{Instance-level comparison between the runner-up and Project2Task routing strategies. Means and standard deviations are computed over five repeated evaluations.}
\label{tab:instance_ablation}
\end{center}

To examine whether this gain holds across individual project queries, we report per-query average scores in Table~\ref{tab:instance_ablation}. The graph-guided router outperforms the runner-up strategy in 9 out of 10 queries. The only exception is \texttt{query1}, where the runner-up strategy obtains a slightly higher score than our selected strategy (6.84 vs. 6.68). One possible explanation is that \texttt{query1} admits multiple plausible decomposition structures, making the routing decision less clear-cut. For the remaining queries, the selected strategy leads to higher portfolio-quality scores, with larger margins on \texttt{query6} (+1.08) and \texttt{query9} (+1.32). These results suggest that graph-guided strategy routing is an important component for turning broad project briefs into well-structured task portfolios.

\begin{center}
\small
\begin{tabular}{@{}ccc@{}}
\hline
\textbf{Query} & \textbf{AutoResearchClaw} & \textbf{Project2Task} \\
\hline
1  & 0.000 & 0.650 \\
2  & 0.856 & 0.885 \\
3  & 0.774 & 0.859 \\
4  & 0.000 & 0.277 \\
5  & 0.699 & 0.832 \\
6  & 0.747 & 0.542 \\
7  & 0.823 & 0.978 \\
8  & 0.975 & 0.954 \\
9  & 0.487 & 0.760 \\
10 & 0.000 & 0.853 \\
\hline
\textbf{Avg. (All)}  & \textbf{0.536} & \textbf{0.759} \\
\textbf{Avg. (Excl. Failures)} & \textbf{0.766} & \textbf{0.830} \\
\hline
\end{tabular}
\captionof{table}{Engineering capability evaluation: task completion accuracy comparing original AutoResearchClaw baseline with our project-level planning method. Higher scores indicate better task performance. A score of 0.000 indicates invalid execution with no valid evaluation result. Avg. (All) includes all tasks; Avg. (Excl. Failures) excludes invalid-execution cases.}
\label{tab:engineering_results}
\end{center}

\subsubsection{Downstream Execution:}

Table~\ref{tab:engineering_results} summarizes task accuracy across the 10 concrete tasks. The average accuracy increases from 0.536 to 0.759 when the executor receives project-level planning contracts. A score of 0.000 denotes invalid execution rather than a measured zero test accuracy: in these cases, the agent-generated experimental design failed during execution and did not produce a valid evaluation result. Because the tasks come from different datasets and domains, this average should be read as a coarse summary rather than a directly comparable metric across all tasks. At the task level, our method produces higher accuracy on 8 out of 10 tasks. Even when excluding the three invalid-execution cases, the average remains higher for our method (0.830 vs. 0.766), suggesting that structured planning can provide useful execution context beyond simply avoiding total failures.

\paragraph{Analysis}
The observed gains appear to come from two factors: \textbf{Precise Context Guidance}, which replaces ambiguous prompts with detailed \textit{handoff contracts} specifying success criteria and boundary constraints; and \textbf{Structured Decomposition}, which breaks down complex projects into \textit{innovation atoms} with clearer boundaries and validation paths. A concrete example is provided in Appendix~\ref{app:contract_example}.

\paragraph{Failure Cases.}
Despite overall gains, performance is not uniform; notably, Query 6 declines from 0.747 to 0.542. This highlights a limitation in our decoupled design: while the planner provides useful decomposition, downstream execution remains bounded by the executor's capabilities. Specifically, the handoff contract may inadvertently \textit{over-constrain} the search space or misalign with the executor's code-generation strengths. Since structural guidance alone cannot overcome execution bottlenecks, future iterations should \textit{co-optimize} both layers, enabling feedback from the executor to refine the planner's contracts.

\section{Conclusion}
We introduced \method{}, a project-level planning layer for autonomous research systems. The work addresses a gap left by autoresearch agents operating on a single task: broad research briefs often contain multiple objectives, shared resources, benchmark requirements, analyses, and staged dependencies that cannot be reliably handled as one monolithic task or as independent task prompts. \method{} makes this pre-execution structure explicit by building an innovation-atom lineage graph, routing it through horizontal, vertical, and hybrid decomposition templates, and exporting dependency-aware task contracts with contribution ownership, shared assets, boundaries, evaluation requirements, and execution order. Our results show that this planning layer improves task-portfolio quality, as measured through generated manuscripts, relative to brief-level and topic-only baselines, and that its structured contracts can improve executor performance when used as task context. These findings suggest that autonomous research systems should treat project-to-task planning as a first-class step before downstream task execution, especially when the goal is a coherent, non-redundant, and executable portfolio of autoresearch tasks.

\bibliography{aaai2027}

@article{lu2024aiscientist,
    author  = {Chris Lu and Cong Lu and Robert Tjarko Lange and Jakob Foerster and Jeff Clune and David Ha},
    title   = {The {AI} Scientist: Towards Fully Automated Open-Ended Scientific Discovery},
    journal = {Computing Research Repository},
    volume  = {arXiv:2408.06292},
    year    = {2024},
    url     = {https://arxiv.org/abs/2408.06292}
}

@article{schmidgall2025agentlaboratory,
    author  = {Samuel Schmidgall and Yusheng Su and Ze Wang and Ximeng Sun and Jialian Wu and Xiaodong Yu and Jiang Liu and Michael Moor and Zicheng Liu and Emad Barsoum},
    title   = {Agent Laboratory: Using {LLM} Agents as Research Assistants},
    journal = {Computing Research Repository},
    volume  = {arXiv:2501.04227},
    year    = {2025},
    url     = {https://arxiv.org/abs/2501.04227}
}

@misc{karpathy2022nanogpt,
    author       = {Andrej Karpathy},
    title        = {{nanoGPT}: The Simplest, Fastest Repository for Training/Finetuning Medium-Sized {GPT}s},
    year         = {2022},
    howpublished = {\url{https://github.com/karpathy/nanoGPT}},
    note         = {Software repository}
}

@article{qu2026bilevel,
    author  = {Yaonan Qu and Meng Lu},
    title   = {Bilevel Autoresearch: Meta-Autoresearching Itself},
    journal = {Computing Research Repository},
    volume  = {arXiv:2603.23420},
    year    = {2026},
    url     = {https://arxiv.org/abs/2603.23420}
}

@article{ferreira2026autoresearch,
    author  = {Fabio Ferreira and Lucca Wobbe and Arjun Krishnakumar and Frank Hutter and Arber Zela},
    title   = {Can {LLM}s Beat Classical Hyperparameter Optimization Algorithms? {A} Study on Autoresearch},
    journal = {Computing Research Repository},
    volume  = {arXiv:2603.24647},
    year    = {2026},
    url     = {https://arxiv.org/abs/2603.24647}
}

@article{lala2023paperqa,
    author  = {Jakub L{\'a}la and Odhran O'Donoghue and Aleksandar Shtedritski and Sam Cox and Samuel G. Rodriques and Andrew D. White},
    title   = {{PaperQA}: Retrieval-Augmented Generative Agent for Scientific Research},
    journal = {Computing Research Repository},
    volume  = {arXiv:2312.07559},
    year    = {2023},
    url     = {https://arxiv.org/abs/2312.07559}
}

@article{asai2024openscholar,
    author  = {Akari Asai and Jacqueline He and Rulin Shao and Weijia Shi and Amanpreet Singh and Joseph Chee Chang and Kyle Lo and Luca Soldaini and Sergey Feldman and Mike D'arcy and David Wadden and Matt Latzke and Minyang Tian and Pan Ji and Shengyan Liu and Hao Tong and Bohao Wu and Yanyu Xiong and Luke Zettlemoyer and Graham Neubig and Dan Weld and Doug Downey and Yih, Wen-tau and Pang Wei Koh and Hannaneh Hajishirzi},
    title   = {{OpenScholar}: Synthesizing Scientific Literature with Retrieval-augmented {LM}s},
    journal = {Computing Research Repository},
    volume  = {arXiv:2411.14199},
    year    = {2024},
    url     = {https://arxiv.org/abs/2411.14199}
}

@article{shao2024storm,
    author  = {Yijia Shao and Yucheng Jiang and Theodore A. Kanell and Peter Xu and Omar Khattab and Monica S. Lam},
    title   = {Assisting in Writing Wikipedia-like Articles From Scratch with Large Language Models},
    journal = {Computing Research Repository},
    volume  = {arXiv:2402.14207},
    year    = {2024},
    url     = {https://arxiv.org/abs/2402.14207}
}

@article{yao2022react,
    author  = {Shunyu Yao and Jeffrey Zhao and Dian Yu and Nan Du and Izhak Shafran and Karthik Narasimhan and Yuan Cao},
    title   = {{ReAct}: Synergizing Reasoning and Acting in Language Models},
    journal = {Computing Research Repository},
    volume  = {arXiv:2210.03629},
    year    = {2022},
    url     = {https://arxiv.org/abs/2210.03629}
}

@article{wang2023planandsolve,
    author  = {Lei Wang and Wanyu Xu and Yihuai Lan and Zhiqiang Hu and Yunshi Lan and Roy Ka-Wei Lee and Ee-Peng Lim},
    title   = {Plan-and-Solve Prompting: Improving Zero-Shot Chain-of-Thought Reasoning by Large Language Models},
    journal = {Computing Research Repository},
    volume  = {arXiv:2305.04091},
    year    = {2023},
    url     = {https://arxiv.org/abs/2305.04091}
}

@article{wu2023autogen,
    author  = {Qingyun Wu and Gagan Bansal and Jieyu Zhang and Yiran Wu and Beibin Li and Erkang Zhu and Li Jiang and Xiaoyun Zhang and Shaokun Zhang and Jiale Liu and Ahmed Hassan Awadallah and Ryen W. White and Doug Burger and Chi Wang},
    title   = {{AutoGen}: Enabling Next-Gen {LLM} Applications via Multi-Agent Conversation},
    journal = {Computing Research Repository},
    volume  = {arXiv:2308.08155},
    year    = {2023},
    url     = {https://arxiv.org/abs/2308.08155}
}

@article{li2023camel,
    author  = {Guohao Li and Hasan Abed Al Kader Hammoud and Hani Itani and Dmitrii Khizbullin and Bernard Ghanem},
    title   = {{CAMEL}: Communicative Agents for ``Mind'' Exploration of Large Language Model Society},
    journal = {Computing Research Repository},
    volume  = {arXiv:2303.17760},
    year    = {2023},
    url     = {https://arxiv.org/abs/2303.17760}
}

@article{hong2023metagpt,
    author  = {Sirui Hong and Mingchen Zhuge and Jiaqi Chen and Xiawu Zheng and Yuheng Cheng and Ceyao Zhang and Jinlin Wang and Zili Wang and Steven Ka Shing Yau and Zijuan Lin and Liyang Zhou and Chenyu Ran and Lingfeng Xiao and Chenglin Wu and J{\"u}rgen Schmidhuber},
    title   = {{MetaGPT}: Meta Programming for {A} Multi-Agent Collaborative Framework},
    journal = {Computing Research Repository},
    volume  = {arXiv:2308.00352},
    year    = {2023},
    url     = {https://arxiv.org/abs/2308.00352}
}

@article{huang2023mlagentbench,
    author  = {Qian Huang and Jian Vora and Percy Liang and Jure Leskovec},
    title   = {{MLAgentBench}: Evaluating Language Agents on Machine Learning Experimentation},
    journal = {Computing Research Repository},
    volume  = {arXiv:2310.03302},
    year    = {2023},
    url     = {https://arxiv.org/abs/2310.03302}
}

@article{chen2024scienceagentbench,
    author  = {Ziru Chen and Shijie Chen and Yuting Ning and Qianheng Zhang and Boshi Wang and Botao Yu and Yifei Li and Zeyi Liao and Chen Wei and Zitong Lu and Vishal Dey and Mingyi Xue and Frazier N. Baker and Benjamin Burns and Daniel Adu-Ampratwum and Xuhui Huang and Xia Ning and Song Gao and Yu Su and Huan Sun},
    title   = {{ScienceAgentBench}: Toward Rigorous Assessment of Language Agents for Data-Driven Scientific Discovery},
    journal = {Computing Research Repository},
    volume  = {arXiv:2410.05080},
    year    = {2024},
    url     = {https://arxiv.org/abs/2410.05080}
}

@article{starace2025paperbench,
    author  = {Giulio Starace and Oliver Jaffe and Dane Sherburn and James Aung and Jun Shern Chan and Leon Maksin and Rachel Dias and Evan Mays and Benjamin Kinsella and Wyatt Thompson and Johannes Heidecke and Amelia Glaese and Tejal Patwardhan},
    title   = {{PaperBench}: Evaluating {AI}'s Ability to Replicate {AI} Research},
    journal = {Computing Research Repository},
    volume  = {arXiv:2504.01848},
    year    = {2025},
    url     = {https://arxiv.org/abs/2504.01848}
}

@misc{liu2026autoresearchclawselfreinforcingautonomousresearch,
      title={AutoResearchClaw: Self-Reinforcing Autonomous Research with Human-AI Collaboration}, 
      author={Jiaqi Liu and Shi Qiu and Mairui Li and Bingzhou Li and Haonian Ji and Siwei Han and Xinyu Ye and Peng Xia and Zihan Dong and Meng Chen and Congyu Zhang and Letian Zhang and Guiming Chen and Haoqin Tu and Xinyu Yang and Lu Feng and Xujiang Zhao and Haifeng Chen and Jiawei Zhou and Xiao Wang and Weitong Zhang and Hongtu Zhu and Yun Li and Jieru Mei and Hongliang Fei and Jiaheng Zhang and Linjie Li and Linjun Zhang and Yuyin Zhou and Sheng Wang and Caiming Xiong and James Zou and Zeyu Zheng and Cihang Xie and Mingyu Ding and Huaxiu Yao},
      year={2026},
      eprint={2605.20025},
      archivePrefix={arXiv},
      primaryClass={cs.AI},
      url={https://arxiv.org/abs/2605.20025}, 
}

@misc{xu2026nanoresearchcoevolvingskillsmemory,
      title={NanoResearch: Co-Evolving Skills, Memory, and Policy for Personalized Research Automation}, 
      author={Jinhang Xu and Qiyuan Zhu and Yujun Wu and Zirui Wang and Dongxu Zhang and Marcia Tian and Yiling Duan and Siyuan Li and Jingxuan Wei and Sirui Han and Yike Guo and Odin Zhang and Conghui He and Cheng Tan},
      year={2026},
      eprint={2605.10813},
      archivePrefix={arXiv},
      primaryClass={cs.AI},
      url={https://arxiv.org/abs/2605.10813}, 
}

@article{holland1983stochastic,
  title={Stochastic Blockmodels: First Steps},
  author={Holland, Paul W. and Laskey, Kathryn Blackmond and Leinhardt, Samuel},
  journal={Social Networks},
  volume={5},
  number={2},
  pages={109--137},
  year={1983},
  doi={10.1016/0378-8733(83)90021-7}
}

@article{karrer2011stochastic,
  title={Stochastic Blockmodels and Community Structure in Networks},
  author={Karrer, Brian and Newman, M. E. J.},
  journal={Physical Review E},
  volume={83},
  number={1},
  pages={016107},
  year={2011},
  doi={10.1103/PhysRevE.83.016107},
  eprint={1008.3926},
  archivePrefix={arXiv}
}

@inproceedings{qiu-etal-2025-completing,
  title     = {Completing A Systematic Review in Hours instead of Months with Interactive {AI} Agents},
  author    = {Qiu, Rui and Chen, Shijie and Su, Yu and Yen, Po-Yin and Shen, Han Wei},
  booktitle = {Proceedings of the 63rd Annual Meeting of the Association for Computational Linguistics (Volume 1: Long Papers)},
  pages     = {31559--31593},
  year      = {2025},
  doi       = {10.18653/v1/2025.acl-long.1523},
  url       = {https://aclanthology.org/2025.acl-long.1523/}
}

@inproceedings{ma-etal-2026-intragent,
  title     = {{I}ntr{A}gent: An {LLM} Agent for Content-Grounded Information Retrieval through Literature Review},
  author    = {Ma, Fengbo and Rao, Zixin and Li, Xiaoting and Chen, Zhetao and Sun, Hongyue and Zhao, Yiping and Chen, Xianyan and Xiang, Zhen},
  booktitle = {Proceedings of the 64th Annual Meeting of the {A}ssociation for {C}omputational {L}inguistics (Volume 1: Long Papers)},
  pages     = {674--715},
  year      = {2026},
  doi       = {10.18653/v1/2026.acl-long.29},
  url       = {https://aclanthology.org/2026.acl-long.29/}
}

@inproceedings{zhang2025aflow,
  author    = {Zhang, Jiayi and Xiang, Jinyu and Yu, Zhaoyang and Teng, Fengwei and Chen, Xiong-Hui and Chen, Jiaqi and Zhuge, Mingchen and Cheng, Xin and Hong, Sirui and Wang, Jinlin and Zheng, Bingnan and Liu, Bang and Luo, Yuyu and Wu, Chenglin},
  title     = {{AFlow}: Automating Agentic Workflow Generation},
  booktitle = {International Conference on Learning Representations},
  pages     = {34040--34077},
  year      = {2025},
  url       = {https://proceedings.iclr.cc/paper_files/paper/2025/file/5492ecbce4439401798dcd2c90be94cd-Paper-Conference.pdf}
}

@inproceedings{ahn2026orchestrationbench,
  author    = {Ahn, Aelim and Lee, Sooyeon and Wang, Hyosun and Park, Chiwan and Kim, Daeryong and Roh, Jihyeon and Yang, Kichang and Jang, Wonjun and {Hwang Woosung} and Kim, Min Seok and Kang, Jihoon},
  title     = {{OrchestrationBench}: {LLM}-Driven Agentic Planning and Tool Use in Multi-Domain Scenarios},
  booktitle = {International Conference on Learning Representations},
  pages     = {115573--115599},
  year      = {2026},
  url       = {https://proceedings.iclr.cc/paper_files/paper/2026/file/bbf38332580c1bed99fa99bc9ee53229-Paper-Conference.pdf}
}

@inproceedings{chen2025mlrbench,
  author    = {Chen, Hui and Xiong, Miao and Lu, Yujie and Han, Wei and Deng, Ailin and He, Yufei and Wu, Jiaying and Li, Yibo and Liu, Yue and Hooi, Bryan},
  title     = {{MLR}-Bench: Evaluating {AI} Agents on Open-Ended Machine Learning Research},
  booktitle = {Advances in Neural Information Processing Systems},
  editor    = {Belgrave, Danielle and Zhang, Cheng and Lin, Hongyu and Pascanu, Razvan and Koniusz, Piotr and Ghassemi, Marzyeh and Chen, Nan},
  publisher = {Curran Associates, Inc.},
  volume    = {38},
  year      = {2025},
  url       = {https://proceedings.neurips.cc/paper_files/paper/2025/file/ab8dd000d6f87f40061a73f8bca7fae4-Paper-Datasets_and_Benchmarks_Track.pdf}
}

@misc{garikaparthi2026researchgym,
  title        = {{ResearchGym}: Evaluating Language Model Agents on Real-World {AI} Research},
  author       = {Aniketh Garikaparthi and Manasi Patwardhan and Arman Cohan},
  year         = {2026},
  eprint       = {2602.15112},
  archivePrefix = {arXiv},
  primaryClass = {cs.AI},
  doi          = {10.48550/arXiv.2602.15112},
  url          = {https://arxiv.org/abs/2602.15112},
  note         = {ICLR 2026 Agents in the Wild Workshop}
}

@misc{qwen36plus,
  author = {{Qwen Team}},
  title  = {{Qwen3.6-Plus}: Towards Real World Agents},
  month  = {April},
  year   = {2026},
  url    = {https://qwen.ai/blog?id=qwen3.6}
}

\appendix
\onecolumn
\raggedbottom 

\section{Task Contract Example}
\label{app:contract_example}

\paragraph{Illustrative Example.}
Consider Query 5 (lightweight time-series classification on UCI HAR). The baseline struggles with abstract constraints such as "lightweight" and "interpretable," often generating overly complex models or exceeding single-GPU limits. Our planner addresses this by translating these concepts into executable specifications (Table~\ref{tab:contract_example}), boosting accuracy from 0.699 to 0.832. Specifically, the contract enforces quantitative boundaries (e.g., $>30\%$ parameter reduction vs. 1D CNNs, strict single-GPU AMP training) and operationalizes abstract goals into verifiable tasks (e.g., mandatory gradient-based attribution stability analysis). By explicitly forbidding multi-GPU or Transformer architectures, it prevents scope drift and ensures the agent focuses on reproducible system design within feasible resource constraints.

\begin{table}[htbp]
\centering
\small
\renewcommand{\arraystretch}{1.3}
\begin{tabular}{@{}p{0.18\linewidth}|p{0.74\linewidth}@{}}
\hline
\textbf{Field} & \textbf{Content} \\
\hline
Must Cover &
Parameter-efficient gated convolution, Gradient-based feature attribution, Single-GPU mixed-precision training, Accuracy-latency Pareto analysis. \\
\hline
Must Not Cover &
Large-scale datasets beyond compact sensors, Multi-GPU/distributed training, Transformer-based or highly specialized external architectures. \\
\hline
Evaluation Boundaries &
\textbf{Data/Hardware:} UCI HAR only; Single-GPU with native PyTorch AMP. \textbf{Metrics:} Parameter count $>30\%$ lower than 1D CNN. \\
\hline
Stage Guidance &
\textbf{Scoping:} Fix UCI HAR preprocessing \& single-GPU VRAM limits. \textbf{Implementation:} Pure PyTorch, modular design, no custom CUDA. \textbf{Analysis:} Compute Pareto frontiers \& validate gradient stability. \\
\hline
\end{tabular}
\caption{Excerpted task contract for Query 5. The contract transforms a vague topic into a scoped, constraint-aware research blueprint.}
\label{tab:contract_example}
\end{table}

\section{Prompt Template for Project-Level Brief Augmentation}
\label{app:prompt_template}

The following prompt template was used to transform single-task research seeds from NanoResearch and ARC-Bench into macro-level project briefs. The prompt enforces constraints on computational feasibility (single-GPU/CPU) and methodological simplicity (no heavy fine-tuning or RAG), while requiring a structured output that supports multi-task autoresearch planning.

The requested sub-directions are candidate content axes used to enrich each brief, rather than a fixed allocation of tasks. The planner may merge or split these directions when determining the size and composition of the task portfolio, and all compared settings receive the same augmented project brief.

\subsection{Role and Task Definition}
\begin{tcolorbox}[
    colback=blue!5!white,
    colframe=blue!75!black,
    colbacktitle=blue!75!black,
    coltitle=white,
    title=\textbf{Role \& Task Definition},
    boxrule=0.5pt,
    arc=3pt,
    auto outer arc,
    boxsep=5pt,
    left=6pt,
    right=6pt,
    breakable,        
    enhanced,  
    top=6pt,
    bottom=6pt,
    width=\linewidth,
    after skip=0pt
]
\small
\textbf{Role:} You are a Senior AI Research Consultant specializing in the systematic expansion and structuring of academic research projects.

\textbf{Task:} Expand the provided \texttt{[Original Research Problem]} into a comprehensive, cohesive \textbf{Project-Level Research Proposal}. The proposed project must contain 3--4 logically coherent and independently verifiable sub-directions that collectively address a unified core scientific question, forming a complete and progressive research narrative.
\end{tcolorbox}

\subsection{Input and Constraints}
\begin{tcolorbox}[
    colback=blue!5!white,
    colframe=blue!75!black,
    colbacktitle=blue!75!black,
    coltitle=white,
    title=\textbf{Input \& Constraints},
    boxrule=0.5pt,
    arc=3pt,
    auto outer arc,
    boxsep=5pt,
    left=6pt,
    right=6pt,
    breakable,         
    enhanced,  
    top=6pt,
    bottom=6pt,
    width=\textwidth,
]
\small
\textbf{Input:} 
\texttt{[Original Research Problem]} (Sourced from the NanoResearch or ARC-Bench datasets)

\textbf{Constraints:}
\begin{itemize}
    \item \textbf{Computational Efficiency:} Maintain a strict "lightweight, efficient, and reproducible" paradigm. All proposed experiments must be strictly feasible on a single GPU or CPU environment, utilizing small batch sizes and optimized memory management.
    \item \textbf{Methodological Focus:} Do \textbf{not} introduce computationally prohibitive or high-cost solutions such as Retrieval-Augmented Generation (RAG), multi-stage continuous pre-training, or full-parameter Large Language Model (LLM) fine-tuning. Prioritize research efforts in three key areas: algorithmic efficiency optimization, lightweight network architectural modifications, and data-centric improvement strategies.
    \item \textbf{Structural Coherence:} The proposed sub-directions must exhibit clear progressive, complementary, or ablation-based relationships. They should logically build upon one another to form a comprehensive and unified research story.
    \item \textbf{Level of Abstraction:} Provide high-level research logic, theoretical motivation, and conceptual frameworks. Do \textbf{not} include low-level implementation details, specific code snippets, or exact hyperparameter configurations.
\end{itemize}
\end{tcolorbox}

\subsection{Output Format and Content Requirements}
\begin{tcolorbox}[
    breakable,         
    enhanced,  
    colback=blue!5!white,
    colframe=blue!75!black,
    colbacktitle=blue!75!black,
    coltitle=white,
    title=\textbf{Output Format \& Requirements},
    boxrule=0.5pt,
    arc=3pt,
    auto outer arc,
    boxsep=5pt,
    left=6pt,
    right=6pt,
    top=6pt,
    bottom=6pt,
    width=\textwidth
]
\small
\textbf{Output Format:} Strictly adhere to the Markdown structure provided below. Do not add any extraneous commentary, introductory text, or concluding remarks outside the defined structure.

\begin{itemize}
    \item \texttt{ [Research Field]}
    \item \texttt{ [Project Title]}
    \item \texttt{ Research Theme}
    \item \texttt{ Background}
    \item \texttt{ Core Problem}
    \item \texttt{ Methodology}
    \item \texttt{ Expected Outcomes}
    \item \texttt{ Significance}
\end{itemize}

\textbf{Content Requirements:}
\begin{enumerate}
    \item \textbf{Research Theme:} Provide a concise, precise statement articulating the overarching scientific inquiry and the primary objective of the project.
    \item \textbf{Background:} Contextualize the research problem within the current landscape of lightweight and efficient AI trends, highlighting the motivation for this specific study.
    \item \textbf{Core Problem:} Clearly define the specific methodological gap, theoretical limitation, or empirical bottleneck present in existing lightweight approaches that this project aims to resolve.
    \item \textbf{Methodology:} Provide a high-level overview of the proposed research methodologies and general investigative directions. Avoid detailing specific experimental protocols; focus strictly on broad, conceptual approaches and the logical flow of the proposed sub-directions.
    \item \textbf{Expected Outcomes:} Describe the anticipated theoretical advancements, empirical contributions, or practical artifacts that will result from the successful execution of this project.
    \item \textbf{Significance:} Highlight the broader impact and specific value of the proposed research for the advancement of resource-constrained, efficient, and accessible AI development.
\end{enumerate}
\end{tcolorbox}

\section{Evaluation Protocol for Manuscript-Based Portfolio Assessment}
\label{app:evaluation_protocol}

This appendix presents the protocol for assessing task-portfolio quality using manuscripts generated by the planned tasks. The protocol covers this output class only; it does not assess other autoresearch artifacts.

\subsection{Evaluator Role and Task Description}

\begin{tcolorbox}[
    colback=blue!5!white,
    colframe=blue!75!black,
    colbacktitle=blue!75!black,
    coltitle=white,
    title=\textbf{Evaluator Role Definition},
    boxrule=0.5pt,
    arc=3pt,
    auto outer arc,
    boxsep=5pt,
    left=6pt,
    right=6pt,
    top=6pt,
    breakable,         
    enhanced,  
    bottom=6pt,
    width=\linewidth,
    after skip=0pt
]
The evaluation is conducted by an AI agent configured with the following system prompt:

\begin{quote}
\textit{You are a senior program committee member of top conferences (NeurIPS/ICLR/AAAI) with exceptional critical thinking and academic review experience. Please evaluate the following collection of papers strictly, objectively, thoroughly, and scientifically.}
\end{quote}
\end{tcolorbox}

\subsection{Input Specifications}

\begin{tcolorbox}[
    colback=blue!5!white,
    colframe=blue!75!black,
    colbacktitle=blue!75!black,
    coltitle=white,
    title=\textbf{Input Specifications},
    boxrule=0.5pt,
    arc=3pt,
    auto outer arc,
    boxsep=5pt,
    left=6pt,
    right=6pt,
    top=6pt,
    bottom=6pt,
    breakable,         
    enhanced,  
    width=\linewidth,
    after skip=0pt
]
The evaluator receives three structured inputs:

\begin{description}[leftmargin=!]
    \item[Project Theme:] The overarching research theme that defines the scope and core scientific question of the project. This serves as the reference standard against which all papers are evaluated for relevance and coverage.
    
    \item[Full Texts of Papers:] The complete text of all papers in the collection, concatenated into a single document. Each paper is prefixed with a stable identifier (\texttt{P1}, \ldots, \texttt{Pk}), and its original section headings are preserved so that rationales can cite evidence unambiguously.
    
    \item[Evaluation Criteria:] The five-dimensional scoring framework detailed in Section~\ref{subsec:scoring_dimensions}.
\end{description}
\end{tcolorbox}

\subsection{Scoring Dimensions and Guidelines}
\label{subsec:scoring_dimensions}

\begin{tcolorbox}[
    colback=blue!5!white,
    colframe=blue!75!black,
    colbacktitle=blue!75!black,
    coltitle=white,
    title=\textbf{Scoring Dimensions and Guidelines},
    boxrule=0.5pt,
    arc=3pt,
    auto outer arc,
    boxsep=5pt,
    left=6pt,
    right=6pt,
    top=6pt,
    bottom=6pt,
    width=\linewidth,
    breakable,          
    enhanced,  
    after skip=0pt
]
\begin{description}[leftmargin=!]
    \item[1. Coherence (1--10 points)] 
    \textbf{Definition:} Measures the logical connection and content cohesion between multiple papers within the same project. \\
    \textbf{Key Questions:}
    \begin{itemize}
        \item Is there a progressive, complementary, supportive, or validating research logic between papers?
        \item Do all papers serve the same overarching research objective?
        \item Is there explicit citation and cross-referencing among papers in the collection?
    \end{itemize}
    \textbf{Scoring Guidelines:}
    \begin{itemize}
        \item \textbf{9--10:} Clear logical relationships, high cross-referencing density, forms a complete integrated whole.
        \item \textbf{7--8:} Natural connections, complete narrative arc, minor details could be optimized.
        \item \textbf{5--6:} Some connection exists but logic is weak; papers are more ``related'' than ``coherent''.
        \item \textbf{3--4:} Only broadly similar topics; lack substantial substantive connections.
        \item \textbf{1--2:} Completely independent; no relationship whatsoever.
    \end{itemize}

    \vspace{0.5em}

    \item[2. Overlap Control (1--10 points)] 
    \textbf{Definition:} Measures the degree of redundant repetition in text, ideas, and core arguments across multiple papers. \\
    \textbf{Key Questions:}
    \begin{itemize}
        \item Are core hypotheses, methods, experimental designs, and conclusions highly similar?
        \item Is there a clear distinction between general foundational content and core innovative content?
        \item Is there extensive textual repetition without proper attribution?
    \end{itemize}
    \textbf{Scoring Guidelines:}
    \begin{itemize}
        \item \textbf{9--10:} Each paper makes independent contributions; almost no overlap in core content.
        \item \textbf{7--8:} Core content is distinct; some overlap in background/methods/experiments is acceptable.
        \item \textbf{5--6:} Different focuses but core problems intersect; uniqueness needs clearer definition.
        \item \textbf{3--4:} Revolve around similar points; core contributions significantly overlap.
        \item \textbf{1--2:} Extensive similarity; core content, methods, data, and conclusions are highly identical.
    \end{itemize}

    \vspace{0.5em}

    \item[3. Coverage (1--10 points)] 
    \textbf{Definition:} Measures the comprehensive coverage of the research topic when considering all papers as a single unit. \\
    \textbf{Key Questions:}
    \begin{itemize}
        \item Is the research content of each paper consistent with the project's research theme?
        \item Does the paper collection cover the key scientific questions, technical dimensions, and application scenarios of the topic?
        \item Does it form a complementary and comprehensive expression without missing important dimensions?
    \end{itemize}
    \textbf{Scoring Guidelines:}
    \begin{itemize}
        \item \textbf{9--10:} Multi-dimensional comprehensive exploration; forms a complete closed loop or three-dimensional picture.
        \item \textbf{7--8:} Covers most important aspects; a few non-core dimensions not addressed.
        \item \textbf{5--6:} Only covers some important dimensions; relatively single perspective with obvious gaps.
        \item \textbf{3--4:} Highly focused on one specific sub-dimension; other aspects completely unaddressed.
        \item \textbf{1--2:} Irrelevant to the topic; severely deviates from the core theme.
    \end{itemize}

    \vspace{0.5em}

    \item[4. Consistency (1--10 points)] 
    \textbf{Definition:} Measures the uniformity of key elements such as terminology, concepts, hypotheses, data, and experimental settings across different papers. \\
    \textbf{Key Questions:}
    \begin{itemize}
        \item Are definitions and usage of core terminology/concepts consistent throughout?
        \item Are basic theoretical assumptions and preconditions consistent?
        \item Are data processing pipelines and evaluation metrics unified? Are there conflicting conclusions?
    \end{itemize}
    \textbf{Scoring Guidelines:}
    \begin{itemize}
        \item \textbf{9--10:} All key elements strictly unified; conclusions mutually support or are fully compatible.
        \item \textbf{7--8:} Core elements consistent; minor non-substantial expression differences exist.
        \item \textbf{5--6:} Individual terminology mixed or data processing inconsistent, but no fundamental contradictions.
        \item \textbf{3--4:} Inconsistent key metrics, ambiguous terminology definitions, or shifting definitions across papers.
        \item \textbf{1--2:} Factual errors or logical contradictions; chaotic experimental settings make results incomparable.
    \end{itemize}

    \vspace{0.5em}

    \item[5. Task Division (1--10 points)] 
    \textbf{Definition:} Measures the rationality of decomposing the macro research topic into sub-projects/papers. \\
    \textbf{Key Questions:}
    \begin{itemize}
        \item Does the sub-project division follow clear and reasonable logic (e.g., research stages, problem levels, technical modules)?
        \item Are sub-project boundaries clear? Is the workload and depth balanced across papers?
        \item Is there excessive workload in some papers or over-concentration of core innovations?
    \end{itemize}
    \textbf{Scoring Guidelines:}
    \begin{itemize}
        \item \textbf{9--10:} Rigorous decomposition logic, clear boundaries, balanced volume, maximizes research efficiency.
        \item \textbf{7--8:} Reasonable decomposition, basically clear division, generally balanced; minor imbalances do not affect the whole.
        \item \textbf{5--6:} Logically feasible but boundaries vague or task definitions unclear; some overlap or imbalance.
        \item \textbf{3--4:} Unreasonable allocation; core innovations concentrated in one paper, others are repetitive/marginal work.
        \item \textbf{1--2:} No logical division or no decomposition at all; chaotic tasks that cannot form a coherent whole.
    \end{itemize}
\end{description}
\end{tcolorbox}

\subsection{Output Requirements}

The evaluator must return exactly one JSON object. Each \texttt{score}
must be an integer in $[1,10]$, and each \texttt{rationale} must identify
at least one paper by its stable identifier and section heading, followed
by concrete evidence. The five dimension names must match those used in
the main text. No additional keys, prose, or Markdown fences are permitted.
\texttt{total\_score} is the arithmetic mean of the five component scores,
rounded to one decimal place. The following valid JSON instance illustrates
the required structure; its values are illustrative.

\begin{tcblisting}{
    enhanced jigsaw,
    breakable,
    listing only,
    colback=gray!5,
    colframe=gray!40,
    arc=2pt,
    boxrule=0.5pt,
    left=3pt,
    right=3pt,
    top=3pt,
    bottom=3pt,
    width=\linewidth,
    listing options={
        basicstyle=\ttfamily\footnotesize,
        breaklines=true,
        columns=fullflexible,
        keepspaces=true,
        showstringspaces=false
    }
}
{
  "project_overview": "The portfolio is coherent and broad, with minor consistency gaps.",
  "scores": {
    "Coherence": {"score": 8, "rationale": "P1 Introduction and P2 Methods state a shared objective and a sequential dependency."},
    "Overlap Control": {"score": 7, "rationale": "P1 Methods and P2 Methods reuse the dataset but define distinct core algorithms."},
    "Coverage": {"score": 8, "rationale": "P1 Results and P2 Results address both project questions; robustness remains unevaluated."},
    "Consistency": {"score": 6, "rationale": "P1 Experimental Setup and P2 Experimental Setup use different data-split conventions."},
    "Task Division": {"score": 7, "rationale": "P1 Contributions and P2 Contributions assign separate development and validation roles."}
  },
  "total_score": 7.2
}
\end{tcblisting}

\end{document}